\documentclass[10pt,twocolumn,letterpaper]{article}

\usepackage{cvpr}
\usepackage{times}
\usepackage{epsfig}
\usepackage{graphicx}
\usepackage{amsmath}
\usepackage{amssymb}
\usepackage{bbm}
\usepackage{chngcntr}

\usepackage{float}
\usepackage{placeins}
\usepackage{graphicx}
\usepackage{subfig}
\usepackage{xcolor}

\newcommand{\thetavec}{\boldsymbol{\theta}}

\newcommand{\Loss}{\mathcal{L}}

\newcommand{\pder}[2]{\frac{\partial#1}{\partial#2}}

\usepackage{soul}
\usepackage{color}

\usepackage[pagebackref=true,breaklinks=true,letterpaper=true,colorlinks,bookmarks=false]{hyperref}

\cvprfinalcopy 

\def\cvprPaperID{****} 

\begin{document}

\title{Lossy Compression with Distortion Constrained Optimization}

\author{Ties van Rozendaal, Guillaume Sauti\`ere, Taco S. Cohen \\
Qualcomm AI Research, Qualcomm Technologies Netherlands B.V. \thanks{Qualcomm AI Research is an initiative of Qualcomm Technologies, Inc. and/or its subsidiaries}\\
{\tt\small \{ties, gsautie, tacos\}@qti.qualcomm.com}
}

\maketitle


\begin{abstract}
When training end-to-end learned models for lossy compression, one has to balance the rate and distortion losses.
This is typically done by manually setting a tradeoff parameter $\beta$, an approach called $\beta$-VAE. 
Using this approach it is difficult to target a specific rate or distortion value, because the result can be very sensitive to $\beta$, and the appropriate value for $\beta$ depends on the model and problem setup.
As a result, model comparison requires extensive per-model $\beta$-tuning, and producing a whole rate-distortion curve (by varying $\beta$) for each model to be compared.

We argue that the constrained optimization method of Rezende  and  Viola, 2018 \cite{Rezende2018-cn} is a lot more appropriate for training lossy compression models because it allows us to obtain the best possible rate subject to a distortion constraint.
This enables pointwise model comparisons, by training two models with the same distortion target and comparing their rate.
We show that the method does manage to satisfy the constraint on a realistic image compression task, outperforms a constrained optimization method based on a hinge-loss, and is more practical to use for model selection than a $\beta$-VAE. 

\end{abstract}

\begin{figure*} 
    \centering
  \subfloat[
    $R$/$D$ performance for constrained optimization and hinge loss baselines. Dashed lines indicate the distortion target value $c_D$.
    \label{fig:dco_vs_hinge}
  ]{%
       \includegraphics[width=0.45\linewidth]{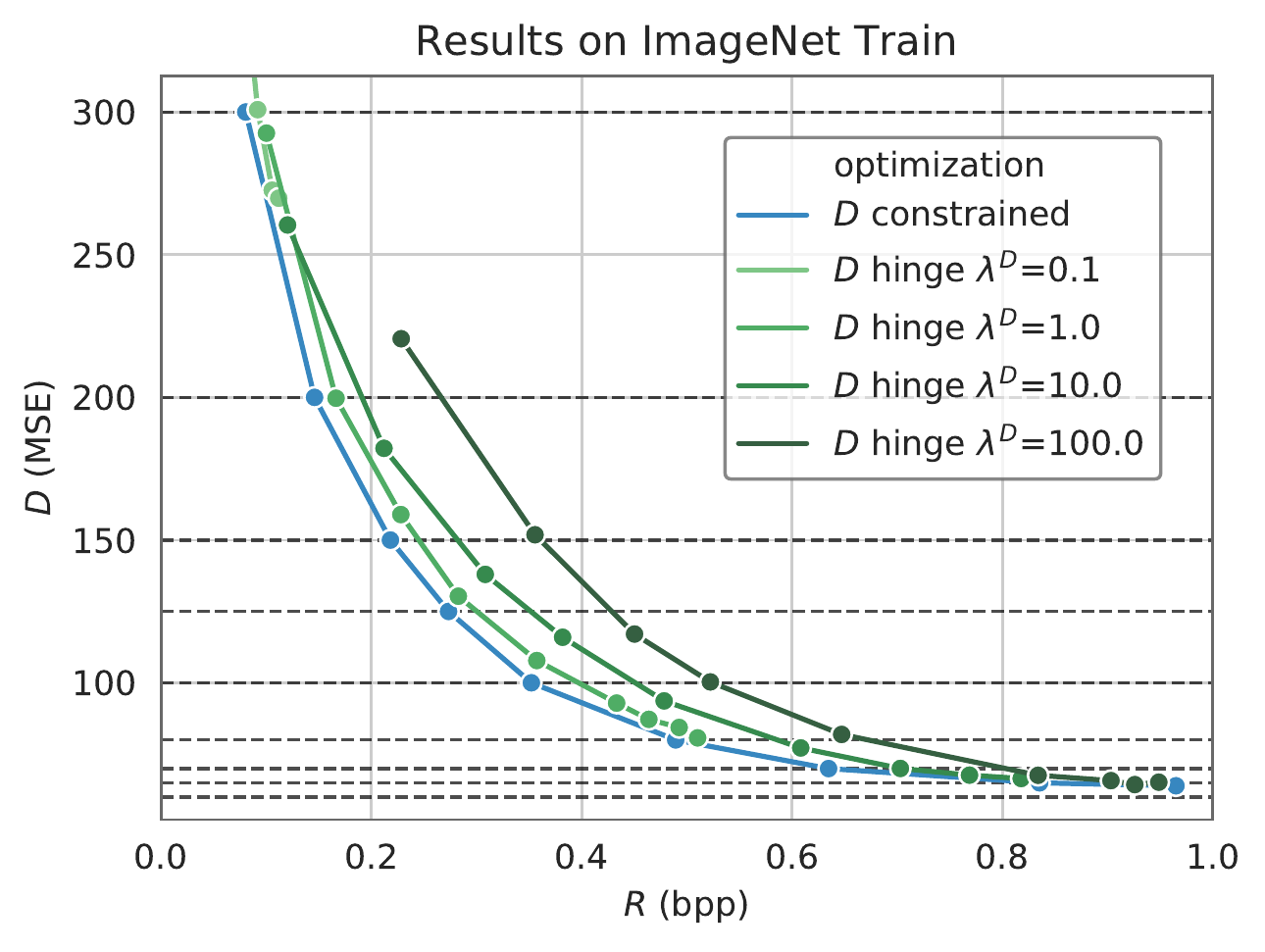}}
    \hfill
  \subfloat[
   $R$/$D$ performance for constrained optimization and $\beta$-VAE.
   \label{fig:dco_vs_beta_vae}
  ]{%
        \includegraphics[width=0.45\linewidth]{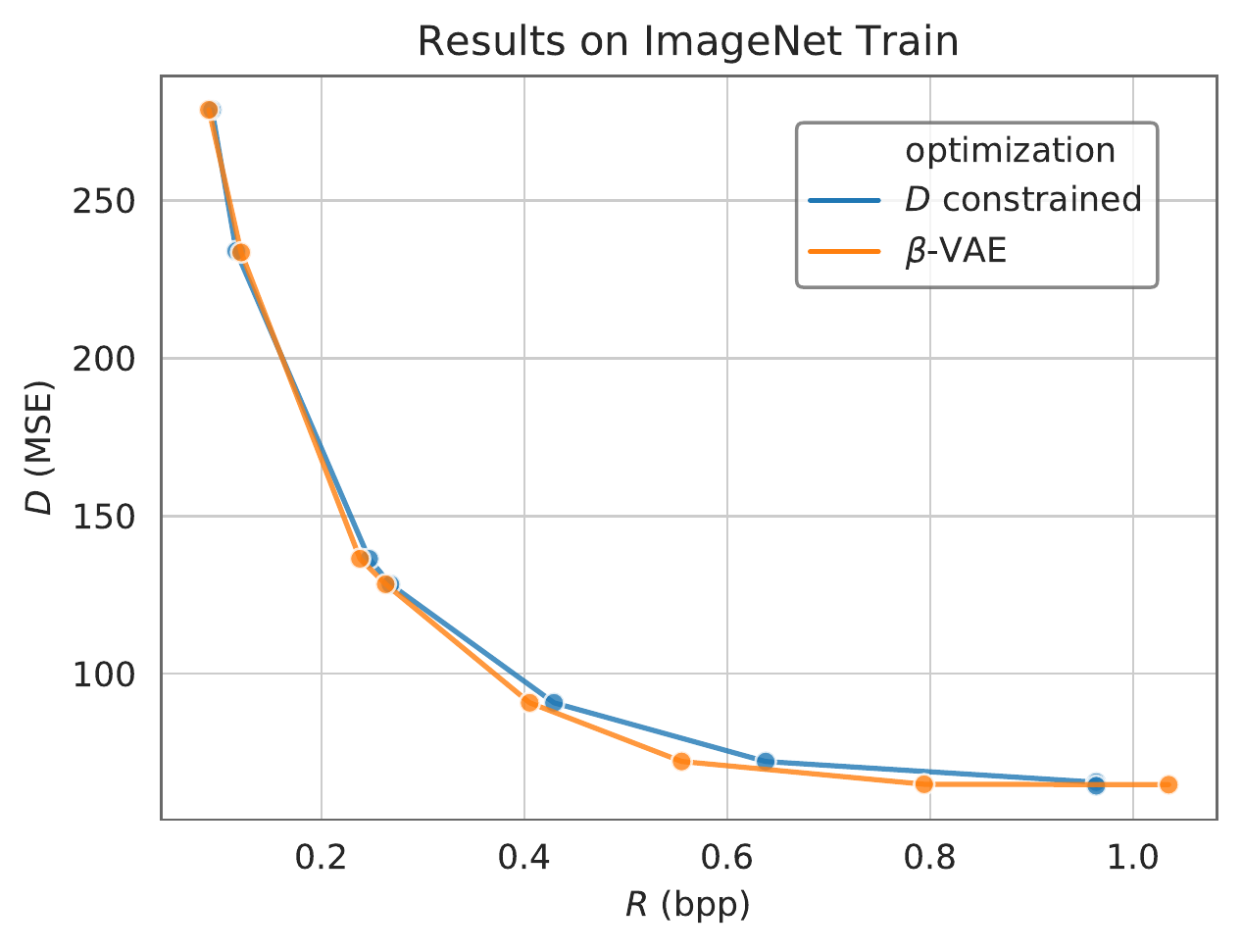}}
  \caption{Training rate / distortion performance for the same model trained using different optimization methods.}
\end{figure*}

\section{Introduction}

Deep latent variable models have started to outperform conventional baselines on lossy compression of images \cite{Cai2019-yl,Choi2019-rd,Minnen_undated-mz,Johnston2019-pb,johnston2018improved,Minnen2018-qt,mentzerConditionalProbabilityModels2018,Theis2017-sl,van2017neural}, video \cite{Liu2019-ey,Djelouah2019-ij,Lu_2019_CVPR,rippel2019learned,Wu2018-bu,NIPS2019_9127,pessoa2018end,Chen2018-wj,Habibian2019-oe}, and audio \cite{Yang2019-ds, van2017neural}.
Nearly all of these methods use a loss function of the form $D + \beta R$, where $D$ measures distortion, $R$ measures bitrate, and $\beta$ is a fixed tradeoff parameter.
We refer to this approach as $\beta$-VAE \cite{higgins2017beta}, because this loss can be motivated from a variational perspective \cite{Habibian2019-oe}.

Despite its popularity, $\beta$-VAE has several drawbacks.
Firstly, setting $\beta$ to target a specific point in the $R$/$D$ plane can be tricky. 
One can show that a model trained with a given $\beta$ should end up at that point on the $R$/$D$ curve where the slope $\pder{R}{D}$ equals $\beta$ \cite{Alemi1938-oi}.
However, because the shape of the $R$/$D$ curve depends on the model and hyperparameters, and because the $R$/$D$ curve can be very steep or flat in the low or high bitrate regime, choosing $\beta$ can be difficult.

Secondly, in order to compare models it is not sufficient to train one instance of each model because the converged models would likely differ in both rate and distortion, which yields inconclusive results unless one model dominates the other on both metrics.
Instead, to compare models we need to train both at several $\beta$ values to generate $R$/$D$ curves that can be compared, which is computationally costly and slows down the research iteration cycle.

A more natural way to target different regions of the $R$/$D$ plane is to set a distortion \emph{constraint} and find our model parameters through constrained optimization:
\begin{align}
\min_{\thetavec} R(\thetavec) \text{ s.t. } D(\thetavec) \leq c_D,
\label{eq:primal_objective}
\end{align}
where $\thetavec$ refers to the joint parameters of the encoder, decoder and prior, and $c_D$ is a distortion target.

We can control the rate-distortion tradeoff by setting the distortion target value $c_D$. Setting this value is more intuitive than setting $\beta$, as it is independent of the slope of the $R$/$D$ curve, and hence independent of model and hyperparameters.

As a result, we can easily compare two different models trained with the same distortion constraint; as we have fixed the $D$ axis we only have to look at the $R$ performance for each model.

Note that one could also minimize the distortion subject to a rate constraint. This is less straightforward as putting too much emphasis on the rate loss at the beginning of training can lead to posterior collapse \cite{bowman-etal-2016-generating,IshaanGulrajani2017,yang2017improved,Raiko2007-ys,So_nderby2016-jr}. 

There is a large literature on constrained optimization, but most of it does not consider stochastic optimization and is limited to convex loss functions.
In this paper we evaluate, in addition to $\beta$-VAE, two constrained optimization methods that are compatible with stochastic gradient descent training of deep networks:
A simple method based on the hinge loss (free bits \cite{Kingma2016-uc,Chen2016-aa,Alemi1938-oi} but applied to distortion rather than rate), and 
the Lagrangian distortion-constrained optimization method of \cite{Rezende2018-cn} ($D$-CO).

We evaluate these methods on a modern image compression system applied to a realistic compression benchmark.
We report on suitable hyperparameters and practical considerations that are relevant in this domain.
We show that $D$-CO outperforms the hinge method, and reaches a similar performance to $\beta$-VAE.
At the same time, $D$-CO is easier to work with and allows for pointwise model selection.

\section{Related Work}

\subsection{Constrained Optimization}
Several works have proposed algorithms to train deep networks under equality or inequality constraints \cite{Marquez-Neila2017-br, Fua2010-ob, Fletcher2013-aq, Rezende2018-cn}.
We deploy the algorithm of \cite{Rezende2018-cn} as the VAE context is most similar to our setup.

The focus of \cite{Rezende2018-cn} is on generative modelling rather than data compression, and there are a number of reasons why the models trained in \cite{Rezende2018-cn} are not directly applicable to data compression.
Firstly, their models contain a stochastic encoder which is not suitable for lossy compression, where bits-back coding is inapplicable.
Secondly, \cite{Rezende2018-cn} do not report $R$/$D$ performance but instead report log-likelihood.
Furthermore, their latent space is continuous while most compression papers use a discrete latent space that allows for entropy coding of the latents under the prior.
Lastly, they use a fixed Guassian prior whereas in lossy compression powerful learnable priors are used to decrease the bitrate as much as possible. 
In this paper we focus on the implementation and evaluation of constrained optimization for practical lossy image compression.

\subsection{Hinge Loss}
Another approach that was proposed for constrained optimization (in the context of avoiding posterior collapse) is free-bits, where the rate loss is hinged $D + \beta \max(R - c_R, 0)$ \cite{Kingma2016-uc,Chen2016-aa,Alemi1938-oi}.
Like constrained optimization, this loss allows us to set a target value, and as such has been used in lossy compression \cite{mentzerConditionalProbabilityModels2018}.
However, we find that this method is inferior to constrained optimization in terms of $R$/$D$ performance and has difficulty converging to the target value.

\subsection{Variable Bitrate Models}
A different approach of dealing with the rate-distortion tradeoff is to train a single model that can compress at different bitrates \cite{todericiVariableRateImage2015, rippel2017real,Choi2019-rd,yang2020variable}. However, some of these works do not meet the performance achieved with specialized models \cite{yang2020variable} or require disjoint training of autoencoder and prior \cite{todericiVariableRateImage2015}. Other methods could  benefit from constrained optimization (e.g. \cite{yang2020variable} still uses $\beta$ multipliers that could be replaced by a distortion target), an exercise that is left for future research.  

\begin{figure*} 
    \centering
  \subfloat[
  \textbf{$D$-CO:} 
  $\min_{\thetavec} R(\thetavec)$ s.t. $D(\thetavec) \leq c_d$
  \label{fig:model_selection_dco_mse}
  ]{%
       \includegraphics[width=0.45\linewidth]{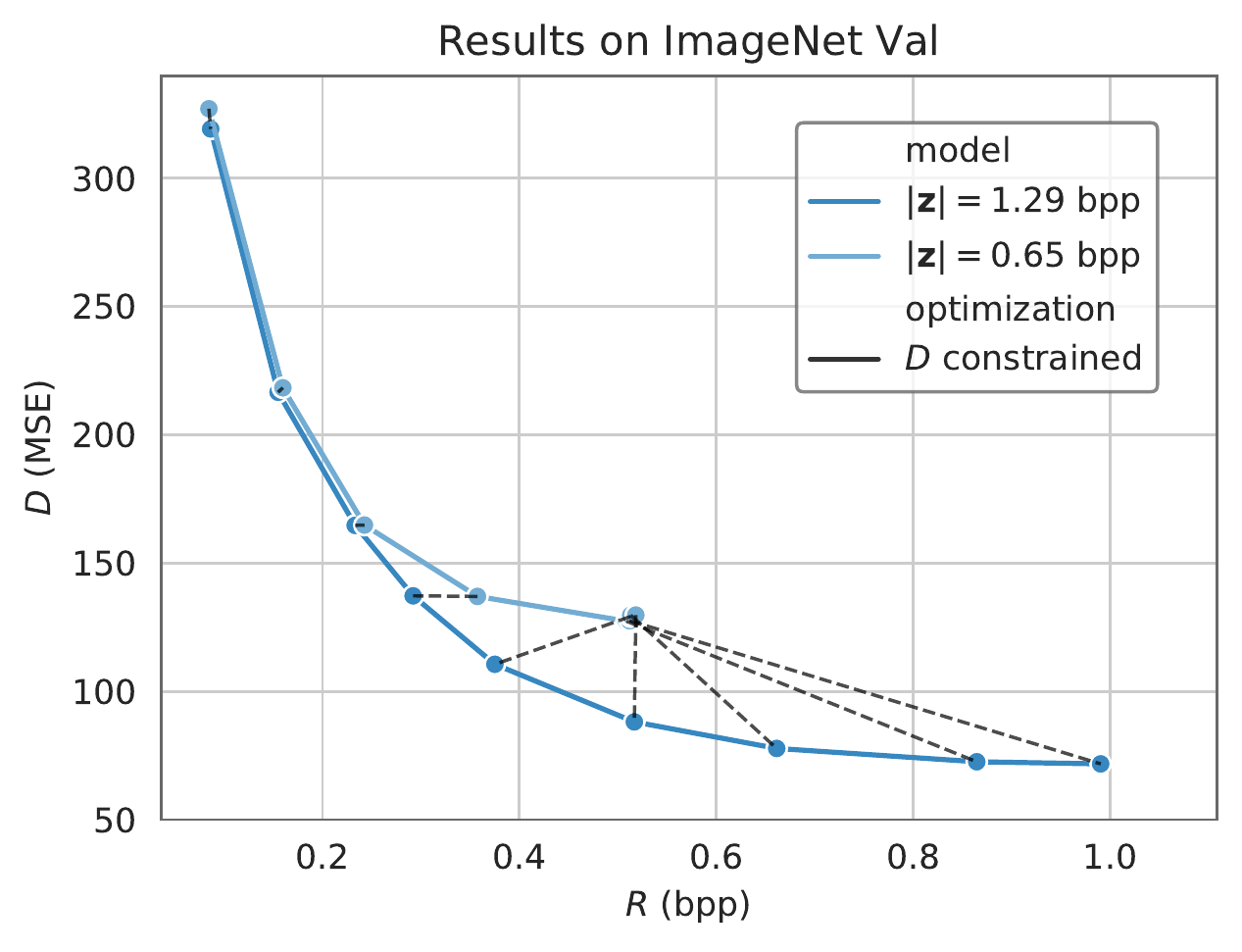}}
    \hfill
  \subfloat[
  \textbf{$\beta$-VAE:} 
  $\min_{\thetavec} D(\thetavec) + \beta R(\thetavec)$
  \label{fig:model_selection_beta_vae_mse}
  ]{%
        \includegraphics[width=0.45\linewidth]{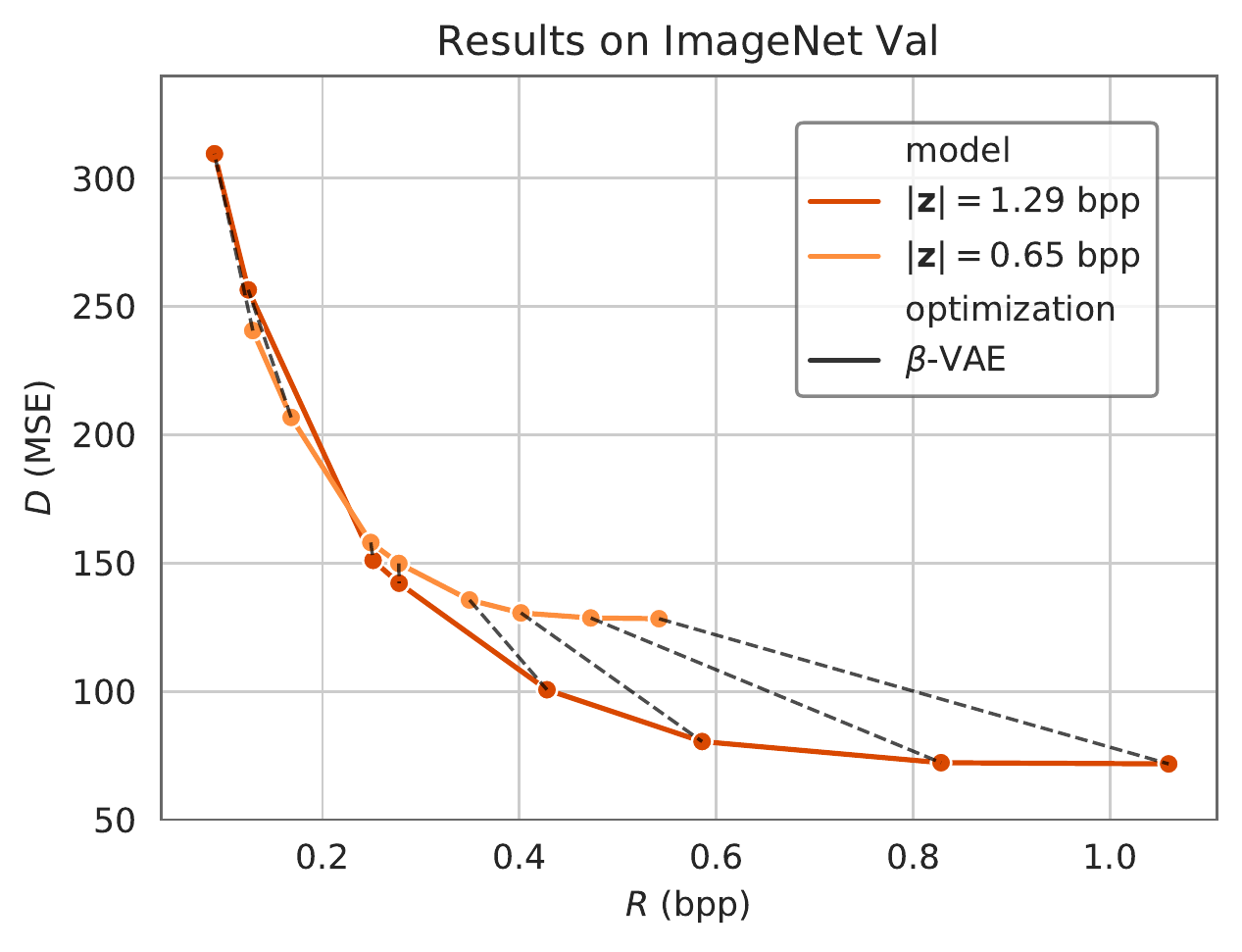}}
  \caption{Model selection using (a) distortion constrained optimization and (b) $\beta$-VAE. The baseline model (dark line) is altered by halving the number of latent channels (light line). Models that are trained using the same distortion target (a) or the same value of  $\beta$ (b) are connected using a dotted line.}
  \label{fig:model_selection}
\end{figure*}

\section{Method}
\subsection{Constrained Optimization}
The Lagrangian of the primal problem in equation \ref{eq:primal_objective} is: 
\begin{align}
\mathcal{L}_\text{Lagrangian} (\thetavec, \lambda^D) = R(\thetavec) + \lambda^D \left ( D(\thetavec) - c_D \right )
\label{eq:loss_lagrangian}
\end{align}

For a convex problem, we would find the minimum of the dual at $\max_{\lambda^D} \min_{\thetavec} \mathcal{L}_\text{Lagrangian}  (\thetavec, \lambda^D)$.

For non-convex deep learning models, we deploy the algorithm proposed by \cite{Rezende2018-cn} and iteratively update $\thetavec$ and  $\lambda^D$ using stochastic gradient descent and ascent respectively.

Note that the $\beta$-VAE loss is the Lagrangian of a rate-constrained optimization. However, the multiplier $\lambda^R = \beta$ is either fixed or updated according to a heuristic schedule \cite{bowman-etal-2016-generating,IshaanGulrajani2017,yang2017improved,Raiko2007-ys,So_nderby2016-jr}, and thus no constrained optimization is performed. 

Because we found that the optimal $D$-CO hyperparameters were different depending on the target value, we normalize our constraint function by the target value. Our loss function thus becomes:

\begin{align}
\mathcal{L}_{D-\text{CO}} (\thetavec, \lambda^D) = R(\thetavec) + \lambda^D \left( \dfrac{D(\thetavec)}{c_D} - 1 \right )
\label{eq:loss_lagrangian_normed}
\end{align}

\subsubsection{Weight and Multiplier Updates}
For each minibatch, we update $\thetavec$ using the Adam optimizer, and $\lambda^D$ using SGD with momentum, to respectively minimize and maximize the batchwise Lagrangian (Eq. \ref{eq:loss_lagrangian_normed}).

Like \cite{Rezende2018-cn}, we reparametrize $\mu^D = \log \lambda^D$ in order to enforce the positivity of $\lambda^D$ (to satisfy the K.K.T. \cite{kuhn2014nonlinear, karush1939minima} conditions for inequality constraints). We also follow them in updating $\Delta \mu^D_{t+1} \propto \pder{\Loss}{\lambda^D_t} = \frac{D_t}{c_D} - 1$ as this resulted in smoother updates of our multipliers than using the actual gradient $(\frac{D_t}{c_D} - 1) \exp{(\mu^D_t)}$.

We use a high momentum ($\alpha=0.99$) for our multiplier updates, to ensure a smooth multiplier trajectory despite the high variance of the MSE loss. As we use the PyTorch \cite{PyTorch} SGD implementation, we make sure to set dampening to be equal to momentum. We clip our log-multiplier s.t. $\lambda^D \leq 10^{3}$ for stability.

\begin{figure*}
 \includegraphics[width=\textwidth]{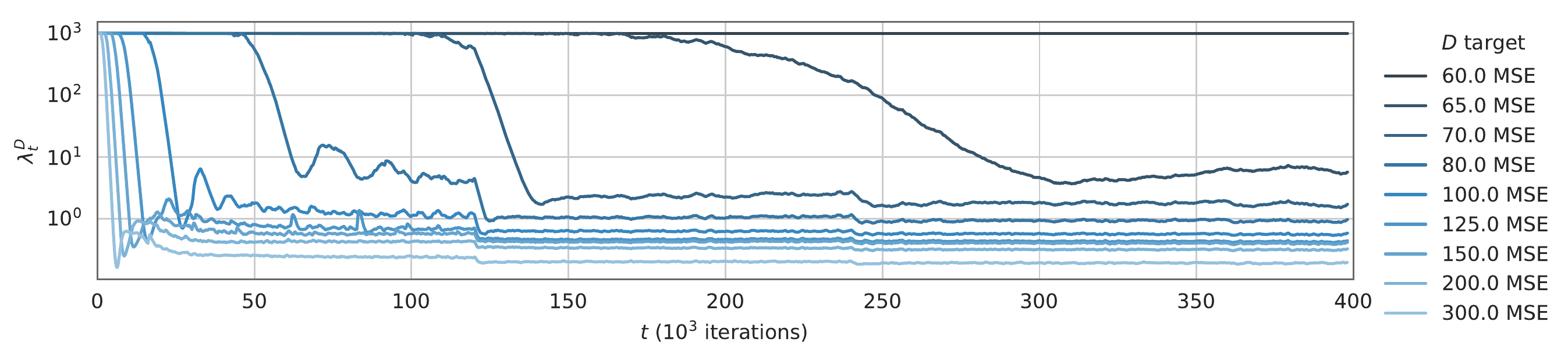}
 \caption{Trajectory of Lagrangian multipliers during training. Each curve represents a model with a different constraint.}
 \label{fig:dco_multipliers}
\end{figure*}

Unlike \cite{Rezende2018-cn} we choose to set our initial value of $\lambda^D$ to the clip value $\lambda^D_0 = 10^{3}$.  This way, we focus on training the autoencoder for distortion at the beginning of training, which we found to be essential for high performance. The final multiplier trajectories are shown in Figure \ref{fig:dco_multipliers}.

\section{Experiments}
We conduct a series of experiments to show how constrained optimization is more suitable for training lossy compression models than $\beta$-VAE or distortion hinge baselines.

\subsection{General Setup}
We use the autoencoder architecture of \cite{mentzerConditionalProbabilityModels2018} but without the mask.
Our prior is the gated pixelCNN \cite{vandenoordConditionalImageGeneration2016} as used in \cite{Habibian2019-oe}. Like \cite{Habibian2019-oe} we jointly train our code-model and autoencoder, without any detaching of the gradients. We use scalar quantization with a learned codebook and a straight-trough estimator (hardmax during forward pass and softmax gradient during backward) \cite{bengioEstimatingPropagatingGradients2013, mentzerConditionalProbabilityModels2018, Habibian2019-oe}.

We train our model on random 160x160 crops of ImageNet Train, and evaluate on 160x160 center crops of ImageNet Validation. Like \cite{mentzerConditionalProbabilityModels2018} we resize the smallest side of all images to 256 to reduce compression artifacts.

We train using the rate loss expressed in bits per pixel (bpp) and using the distortion loss expressed in average MSE computed on unnormalized images on a 0-255 scale.

We update our parameters using Adam with a learning rate of $2 \cdot 10^{-3}$ for the autoencoder and $1 \cdot 10^{-4}$ for the prior. We decay both learning rates every 3 epochs (120087 iterations) by a factor of $0.1$. For the multiplier updates, we use SGD with a learning rate of $5 \cdot 10^{-3}$. We use a batch size of 32.

\subsection{$D$-CO vs. $D$-hinge}
\label{sec:dco_vs_hinge}
For this experiment we choose exponentially spaced $D$-constraint values (60, 65, 70, 80, 100, 125, 150, 200, 300 MSE) and look at how well the methods converge to the set target. We compare our $D$-CO training with the simpler $D$-hinge baselines of the form:
\begin{align}
\mathcal{L}_{D\text{-hinge}} (\thetavec) = R(\thetavec) +  \lambda^D \max \left ( \dfrac{D(\thetavec)}{c_D} - 1 , 0 \right)
\end{align}
Unlike $D$-CO, $\lambda^D$ is fixed during training, but we train models with different values (0.01, 0.1, 1, 10, 100). In line with $D$-CO, we use the normalized constraint function as we verified that it worked better than the unnormalized one.

Results are shown in Figure \ref{fig:dco_vs_hinge}. Observe that the $D$-CO models converge very closely to the set target (within 1 MSE point for achievable constraints). For the hinge models, the constraint is not satisfied reliably and overall $R$/$D$ performance is worse (some models converged to $R$/$D$ values outside of the chosen display range). Furthermore, the hinge models are sensitive to the value of $\lambda_D$, and the optimal value differs per target.

Figure \ref{fig:dco_multipliers} shows the trajectories of the $D$-CO multipliers. For stricter constraints, it takes longer before the multiplier starts to drop, changing emphasis from $D$ to $R$. In the limit of an unachievable constraint (MSE $<62$), the multiplier remains constant at the clip value. All multipliers converge to a relatively stable final value, which is dependent on the target (as expected since the $\pder{R}{D}$ slope is different).

\subsection{$D$-CO vs. $\beta$-VAE}
\label{sec:dco_vs_beta_vae}
In the next experiment, we compare the $R$/$D$ performance of $D$-CO to the $\beta$-VAE baseline. 
We first train $\beta$-VAE models for exponentially spaced $\beta$ values (0.1, 10, 50, 100, 200, 250, 500, 750). For each $\beta$-VAE, we use the distortion loss over the last training epoch as the target for training a $D$-CO model.

Results are shown in Figure \ref{fig:dco_vs_beta_vae} (PSNR results in Figure \ref{fig:dco_beta_vae_psnr}). The $R$/$D$ performance of the $D$-CO models is similar to that of the $\beta$-VAEs. For bitrates higher than 0.4 bpp, we see a slight advantage for the $\beta$-VAE. For these target values, the $D$-CO multipliers are almost constant (see the strict constraints in Figure \ref{fig:dco_multipliers}) and we thus attribute this difference to the optimization hyperparameters being fine-tuned for the scale of the $\beta$-VAE loss.

\subsection{Model Selection}
In the final experiment we highlight how constrained optimization can simplify the model selection process. We adapt our architecture by changing the number of latent channels from $32$ to $16$, effectively halving the maximum channel capacity from 1.29 bpp to 0.64 bpp. We train $\beta$-VAE models for the $\beta$s from Section \ref{sec:dco_vs_beta_vae} and $D$-CO models using the targets from Section \ref{sec:dco_vs_hinge}.

Results are shown in Figure \ref{fig:model_selection}.
For both optimization methods, the lowest achievable distortion has increased from $\sim70 $ MSE to $\sim130 $ MSE for the model with decreased channel capacity.

For the $\beta$-VAE optimization, points with the same $\beta$ now end up at very different points on the $R$/$D$ plane. For the half-capacity model, we cover a narrow range of 240-128 MSE.

In contrast, $D$-CO produces two comparable $R$/$D$ curves. Distortion targets below 130 MSE are unachievable for the half-capacity model and are all collapsed into a single point. However, for any  achievable distortion target, both models end up with a similar distortion which allows us to do a pointwise comparison.

\section{Conclusion}
We present distortion constrained optimization ($D$-CO) as an alternative to $\beta$-VAE training for lossy compression. We report suitable hyperparameters and propose to normalize the constraint function for better performance. We demonstrate that $D$-CO gives similar performance to $\beta$-VAE on a realistic image compression task, while at the same time providing a more intuitive way to balance the rate and distortion losses. Finally, we show how $D$-CO can facilitate the model selection process by allowing pointwise model comparisons. 

{\small
\bibliographystyle{ieee_fullname}
\bibliography{egbib}
}

\clearpage
\appendix
\section{Supplementary Material}
\subsection{PSNR Results}
\FloatBarrier
\counterwithin{figure}{section}
\setcounter{figure}{0}
\begin{figure}[h!] 
 \includegraphics[width=\linewidth]{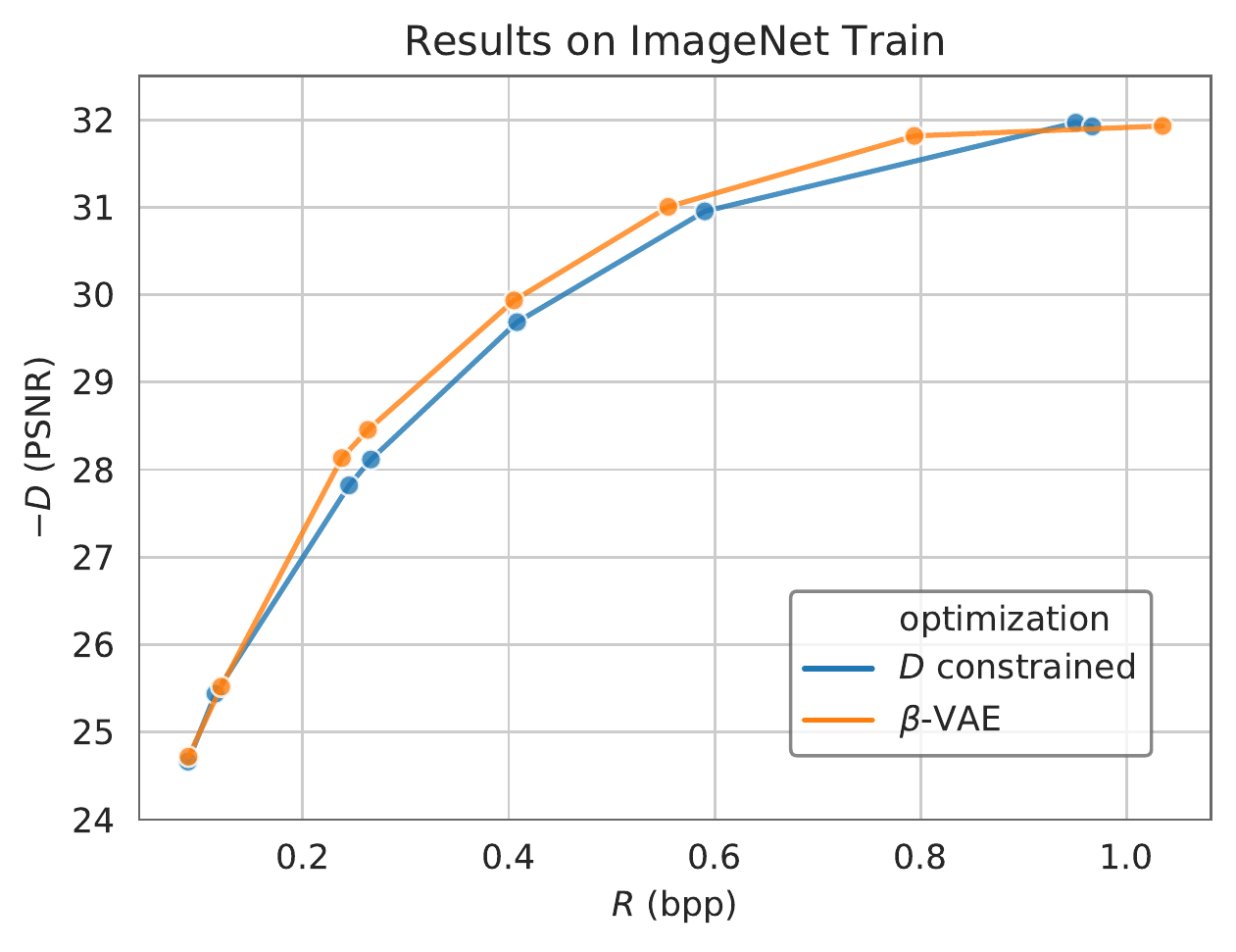}
  \caption{Rate / PSNR performance for $D$ constrained optimization and $\beta$-VAE.}
  \label{fig:dco_beta_vae_psnr}
\end{figure}

\begin{figure}[h!] 
 \includegraphics[width=\linewidth]{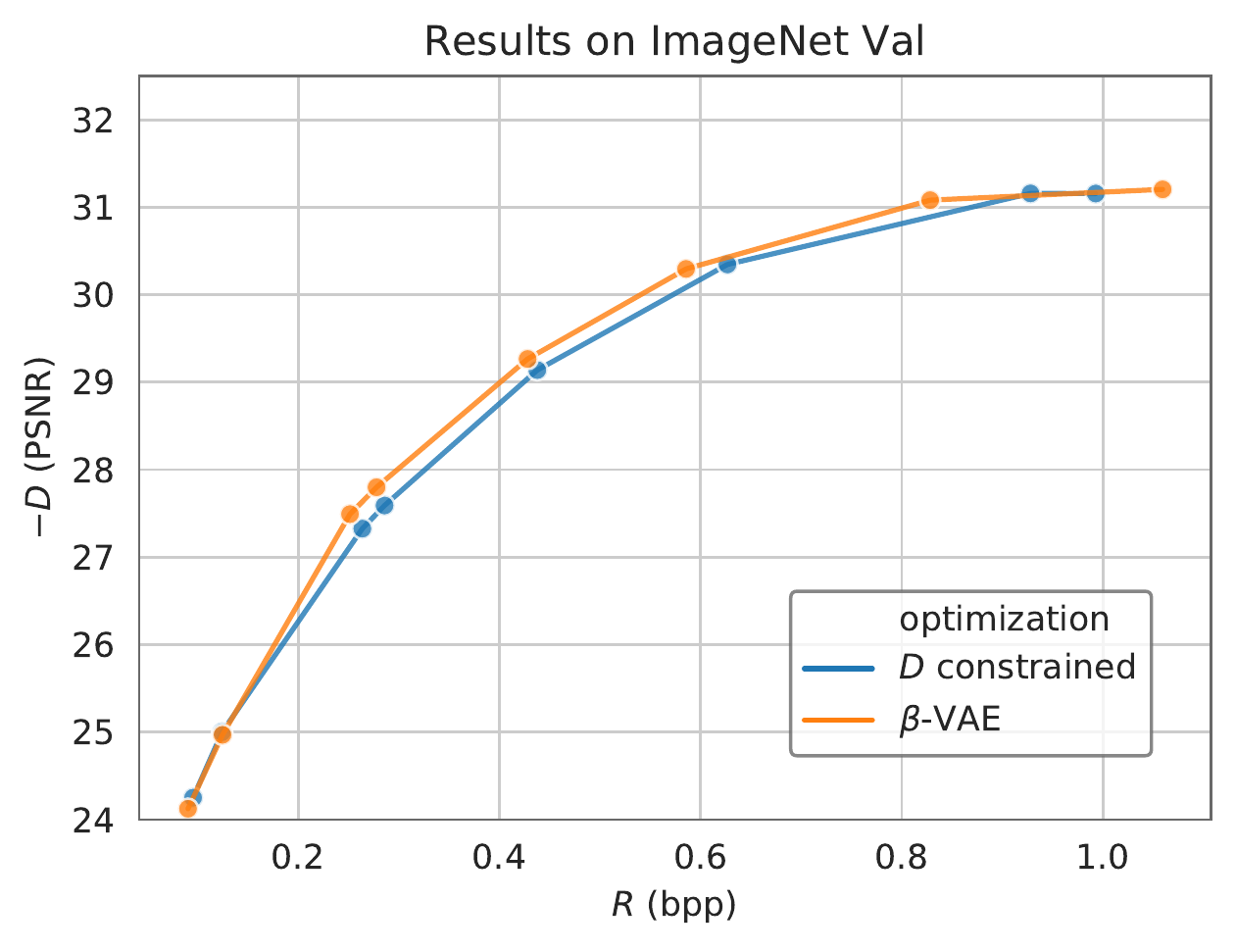}
  \caption{Validation Rate / PSNR performance for $D$ constrained optimization and $\beta$-VAE.}
  \label{fig:dco_beta_vae_psnr_val}
\end{figure}

\end{document}